%% file: main.tex
\documentclass{article} 
\usepackage{iclr2025_conference,times}

\input{math_commands.tex}

\usepackage[hidelinks]{hyperref}
\usepackage{multirow}
\usepackage{booktabs}   
\usepackage{url}
\usepackage{url}
\usepackage{algorithmic}
\usepackage{algorithm}
\usepackage{graphicx}
\usepackage{subcaption}

\title{OneReward: Unified Mask-Guided Image Generation via Multi-Task Human Preference Learning}

\author{Yuan Gong\footnotemark[1] \hspace{0.1cm} 
Xionghui Wang\footnotemark[1]\, \footnotemark[2] \hspace{0.1cm} 
Jie Wu \hspace{0.1cm} Shiyin Wang \hspace{0.1cm} Yitong Wang \hspace{0.1cm} Xinglong Wu \\
\\
ByteDance Inc.\\
}

\iclrfinalcopy 
\makeatletter
\renewcommand{\@oddhead}{}
\renewcommand{\@evenhead}{}
\makeatother

\usepackage{enumitem}
\begin{document}

\maketitle

\footnotetext[1]{Equal contribution.}
\footnotetext[2]{Project leader.}

\input{abstract}

\input{introduction}
\input{relate_work}
\input{preliminaries}

\input{reward}

\input{experiments}
\input{conclusion}

\bibliography{iclr2025_conference}
\bibliographystyle{iclr2025_conference}

\appendix
\input{appendix}

\end{document}

%% file: math_commands.tex
\usepackage{amsmath,amsfonts,bm}

\def\eqref#1{equation~\ref{#1}}

\def\1{\bm{1}}

\DeclareMathAlphabet{\mathsfit}{\encodingdefault}{\sfdefault}{m}{sl}
\SetMathAlphabet{\mathsfit}{bold}{\encodingdefault}{\sfdefault}{bx}{n}



%% file: abstract.tex
\begin{abstract}
In this paper, we introduce OneReward, a unified reinforcement learning framework that enhances the model's generative capabilities across multiple tasks under different evaluation criteria using only \textit{One Reward} model. By employing a single vision-language model (VLM) as the generative reward model, which can distinguish the winner and loser for a given task and a given evaluation criterion, it can be effectively applied to multi-task generation models, particularly in contexts with varied data and diverse task objectives.
We utilize OneReward for mask-guided image generation, which can be further divided into several sub-tasks such as image fill, image extend, object removal, and text rendering, involving a binary mask as the edit area. 
Although these domain-specific tasks share same conditioning paradigm, they differ significantly in underlying data distributions and evaluation metrics.
Existing methods often rely on task-specific supervised fine-tuning (SFT), which limits generalization and training efficiency. 
Building on OneReward, we develop Seedream 3.0 Fill, a mask-guided generation model trained via multi-task reinforcement learning directly on a pre-trained base model, eliminating the need for task-specific SFT.
Experimental results demonstrate that our unified edit model consistently outperforms both commercial and open-source competitors, such as Ideogram, Adobe Photoshop, and FLUX Fill [Pro], across multiple evaluation dimensions.
Code and model are available at: \url{https://one-reward.github.io}
    
\end{abstract}

\begin{figure}[!t]
  \centering

  \begin{minipage}[b]{0.45\textwidth}
    \centering
    \includegraphics[width=\linewidth]{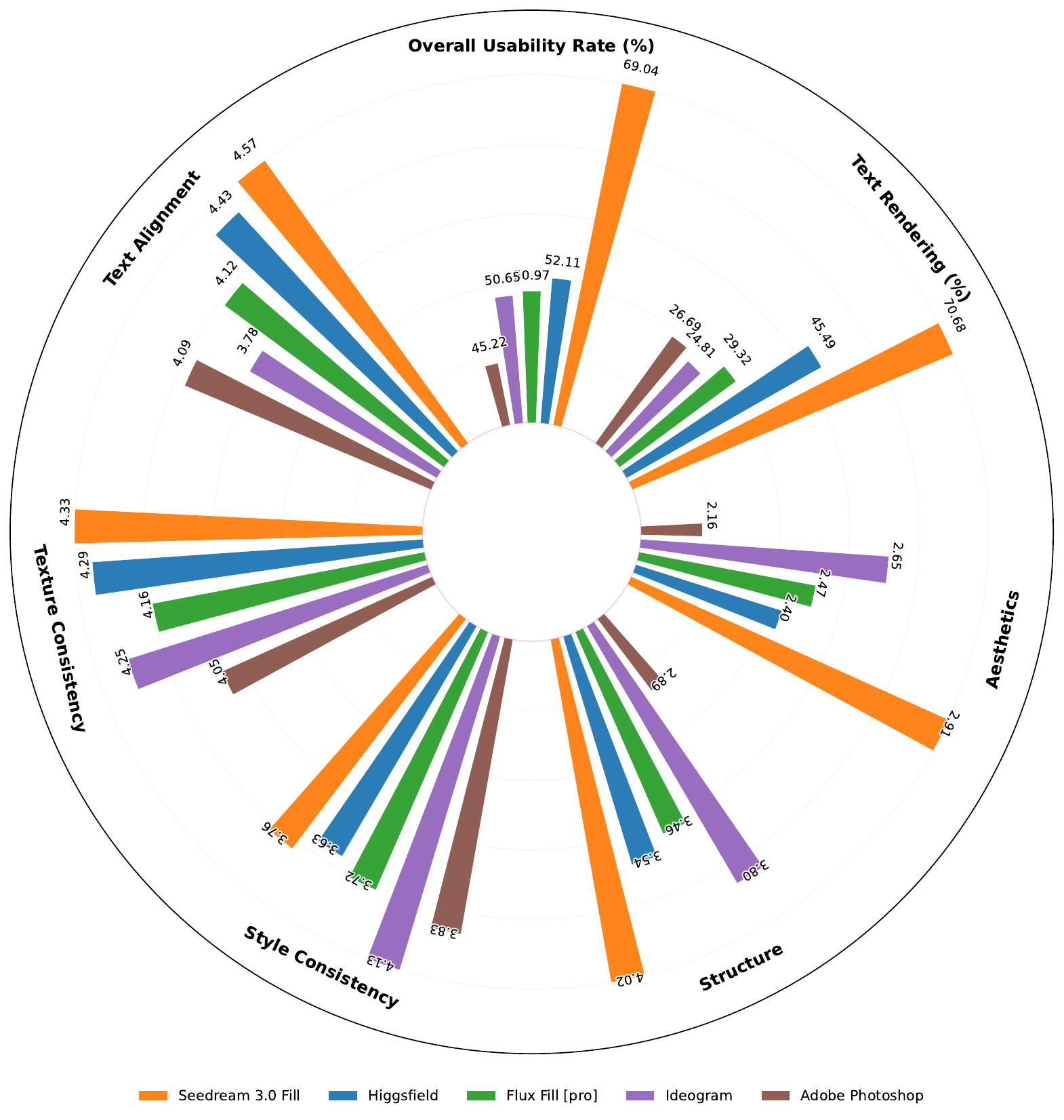}
    \caption*{(a) Image Fill}
  \end{minipage}
  \hspace{0.02\textwidth}
  \begin{minipage}[b]{0.45\textwidth}
    \centering
    \includegraphics[width=\linewidth]{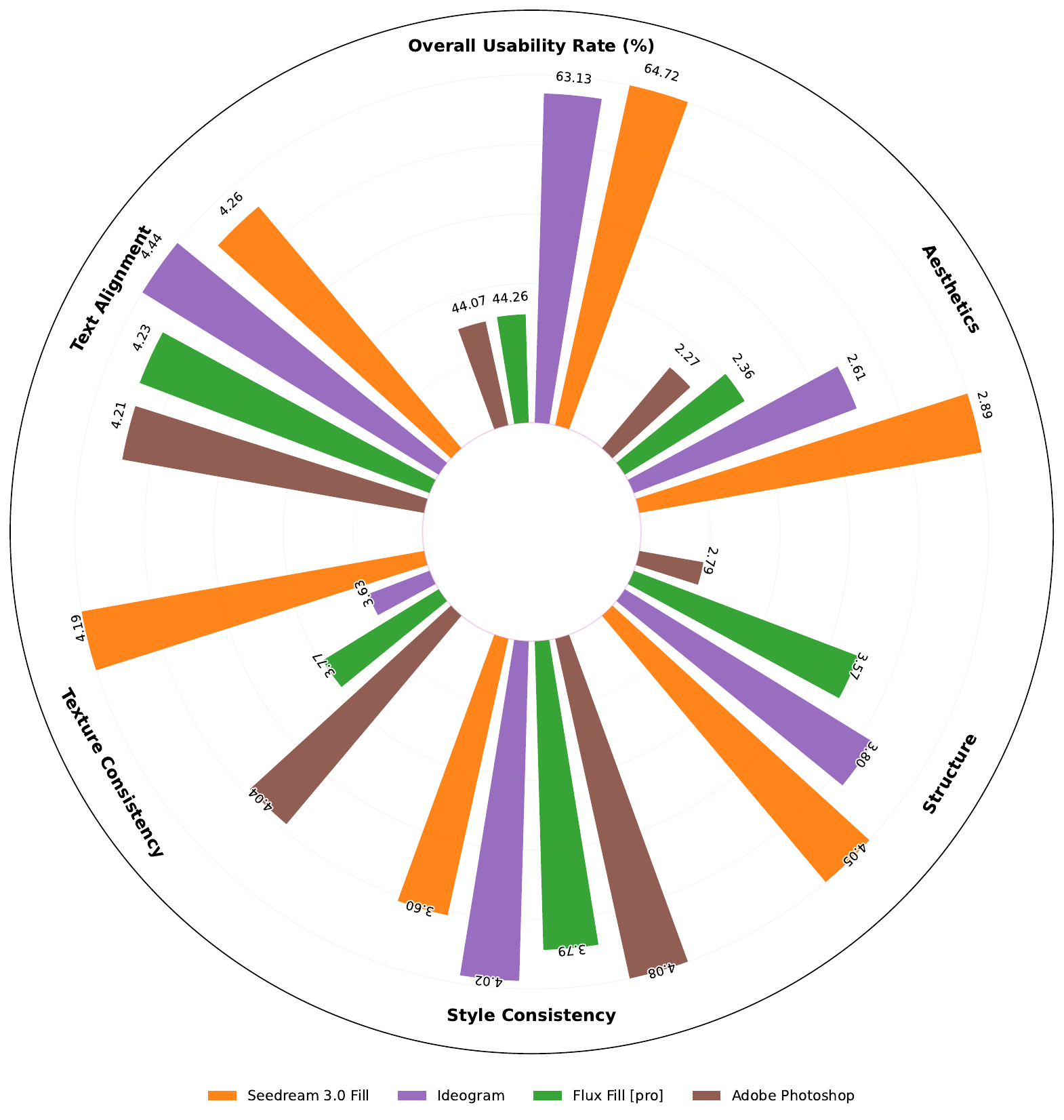}
    \caption*{(b) Image Extend with Prompt}
  \end{minipage}

  \vspace{1em}

  \begin{minipage}[b]{0.45\textwidth}
    \centering
    \includegraphics[width=\linewidth]{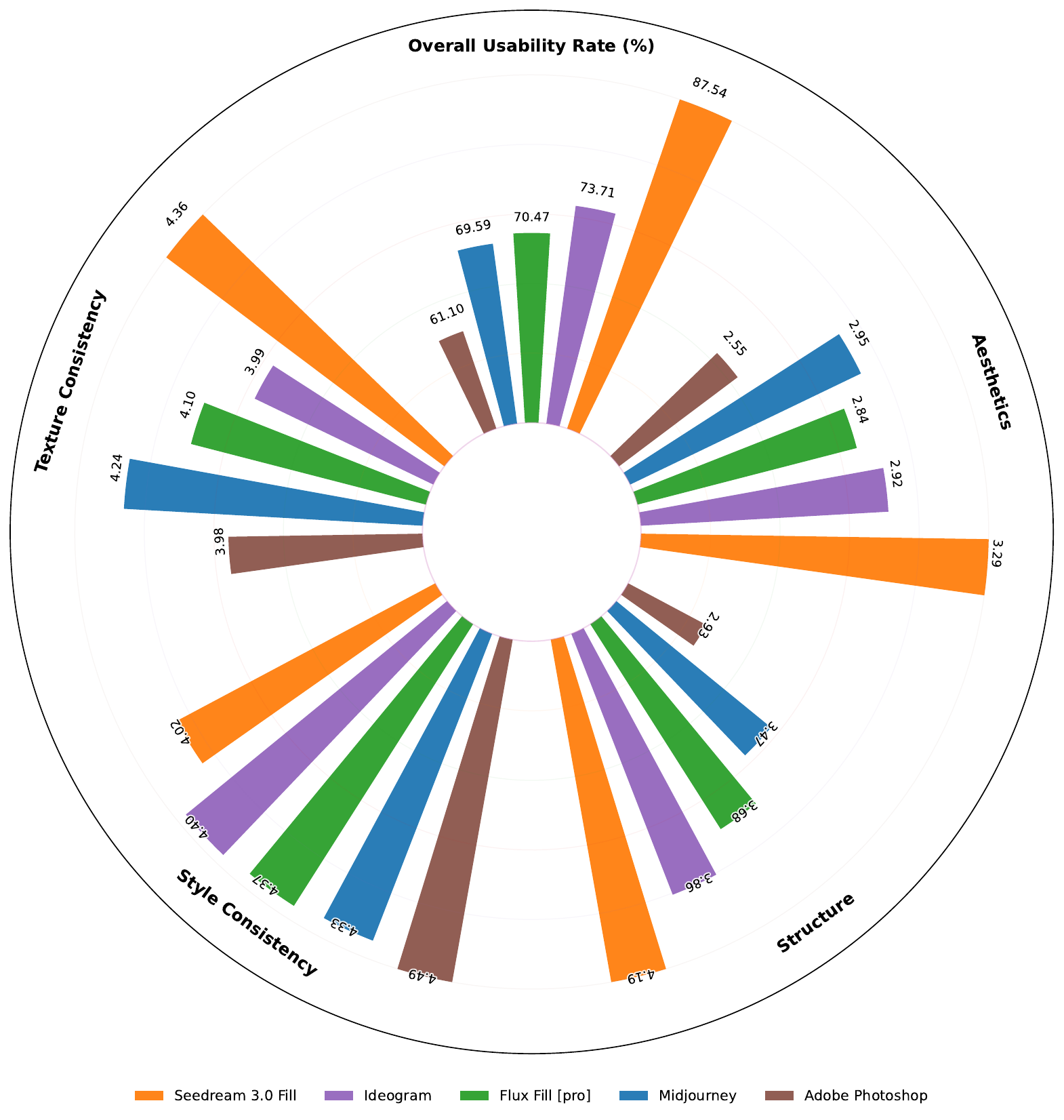}
    \caption*{(c) Image Extend without Prompt}
  \end{minipage}
  \hspace{0.02\textwidth}
  \begin{minipage}[b]{0.45\textwidth}
    \centering
    \includegraphics[width=\linewidth]{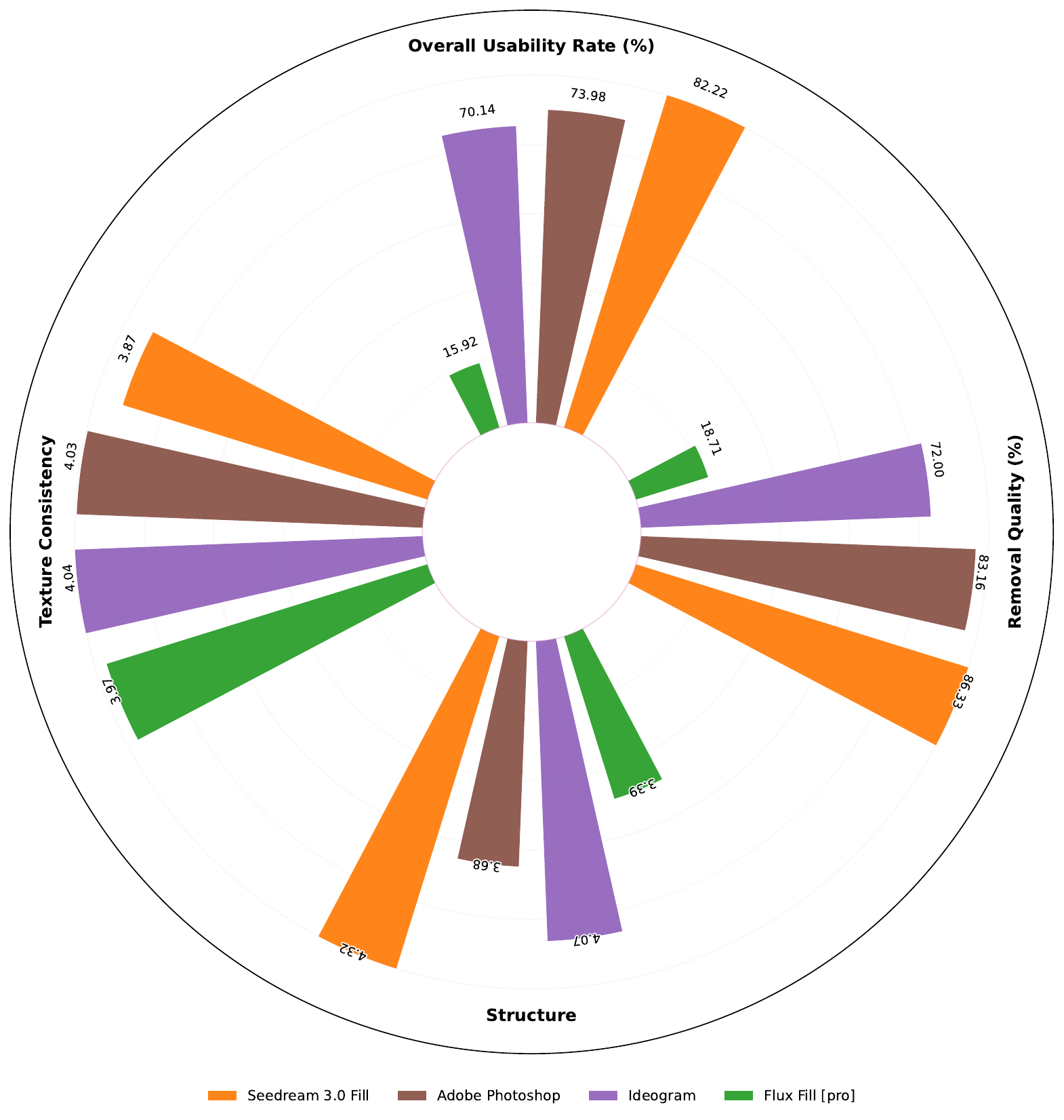}
    \caption*{(d) Object Removal}
  \end{minipage}

  \caption{Overall evaluation across four image editing tasks, and text rendering is included in image fill. For each sub-task, we selected only state-of-the-art models or closed-source APIs as competitors and conducted detailed evaluations in multiple dimensions. Note that different tasks have different evaluation criteria.}
  \label{fig:radar}
\end{figure}

%% file: introduction.tex
\section{Introduction}

Recent advancements in diffusion model(\cite{rombach2022high, podell2023sdxl, flux2024}) have enabled a diverse range of challenging tasks, such as inpainting, outpainting, object removal, and text rendering. Although these tasks share a common mask-guided input format, they exhibit significant divergence in conditional distributions and evaluation metrics, presenting considerable challenge to the development of a unified, versatile model.
Inpainting, also known as image fill, involves modifying or adding specific objects within a localized masked area, emphasizing accurate prompt alignment, aesthetic coherence, and structural integrity. Outpainting, or image-extend, requires generating extensive content around an existing image, expanding beyond its original borders, with a strong emphasis on visual aesthetics, seamless integration, and structural consistency. Object removal entails filling a masked region based on surrounding context, requires avoiding the generation of extra object, and ensuring texture consistency with the original image. Text rendering specifically targets accurate rendering of textual elements, emphasizing precision in generating and aligning fonts according to given instructions.
Current state-of-the-art generative models typically excel within specific editing tasks but struggle to maintain consistently high performance across multiple tasks simultaneously. Existing methods or community models often rely on task-specific supervised fine-tuning (SFT), or LoRA(\cite{hu2022lora}) with limited data base on SD1.5-Inpaint(\cite{rombach2022high}) and FLUX Fill(\cite{flux2024}), which restricts their generalization to diverse editing scenarios. This reveals the difficulty of designing a unified framework capable of supporting multiple image editing tasks while avoiding the inefficiencies of task-specific fine-tuning.

Reinforcement learning from human feedback (RLHF) methods for diffusion and flow matching model, such as Direct Preference Optimization (DPO)(\cite{rafailov2023direct, wallace2024diffusion, xu2024visionreward,liu2025improving}), reward-base method (\cite{xu2023imagereward,zhang2024onlinevpo,li2024controlnet++,gao2025seedance}) and RL-based method (\cite{black2023training,liu2025flow, xue2025dancegrpo}) have shown strong promise in aligning generative outputs with human preferences across text-to-image and text-to-video domains.
However, DPO faces fundamental limitations in handling diverse tasks and evaluation dimensions concurrently, as it inherently assumes a well-defined preference order that may not hold across heterogeneous tasks and criteria.For instance, DPO cannot unambiguously determine the winner and loser when an image is better in aesthetics but worse in structure than its counterpart.
Reward Feedback Learning (ReFL), while significantly boosting model performance in specific dimensions, typically requires training separate reward models for each evaluation criterion when using traditional multimodal architectures such as BLIP(\cite{li2022blip}) and CLIP(\cite{radford2021learning}), increasing training and tuning complexity. Furthermore, ReFL encounters reward conflicts in multi-task scenarios, where high quality object generation may receive completely opposite evaluations in the task of image-fill and object-removal.
FlowGRPO(\cite{liu2025flow}) and DanceGRPO(\cite{xue2025dancegrpo}) introduce GRPO(\cite{shao2024deepseekmath}), which is powerful in Large Language Model(LLM), into flow matching models, by converting deterministic Ordinary Differential Equation(ODE) sampleing into a Stochastic Differential Equation (SDE) framework. While GRPO-based methods significantly enhance performance on vision generation tasks, they rely on policy-based estimation by introducing a group-relative formulation to estimate the advantage, without explicitly maximizing reward signals during optimization. This often results in slower convergence compared to reward-driven approaches.

To overcome these limitations, we introduce OneReward, a unified reinforcement learning framework for mulit-task image generation using only one VLM as the reward model. 
By incorporating task category and evaluation metric information (e.g. aesthetics, structure, consistency) directly into its queries, the VLM can effectively distinguish between tasks and evaluation criteria, enabling it to make pairwise judgments and determine which output is better under certain setting.
Base on OneReward, we adopt Seedream 3.0(\cite{gao2025seedream}) as the pre-trained base model and develop Seedream 3.0 Fill, a state-of-the-art (SOTA) mask-guided image generation model that consistently delivers superior performance across a diverse set of tasks, including image fill, image extension, object removal, and text rendering. Seedream 3.0 Fill is directly optimized via reinforcement learning from a pre-trained model, without any SFT.
During training, we treat the initial pre-trained model as the reference model and the training one as the policy model, optimizing the latter to generate results that surpass the reference model in each task-specific evaluation metric. 
The reward signal is derived from the probability of the token ``Yes'' generated by the VLM, which is then used for gradient backward.
To the best of our knowledge, this is the first work to employ reinforcement learning as a direct optimization paradigm in the context of multi-task image editing.
The main contributions of our work are threefold:
\begin{enumerate}
    \item We propose OneReward, a novel reward model framework for the visual domain by employing VLM as the generative reward model to enhance multi-task reinforcement learning, significantly improving the policy model's generation ability across diverse scenarios.
    \item Building on OneReward, we develop Seedream 3.0 Fill, a unified SOTA image editing model capable of effectively handling diverse tasks including image fill, image extend, object removal, and text rendering. It surpasses several leading commercial and open-source models, including Ideogram, Adobe Photoshop, and FLUX Fill [Pro].
    \item By applying our multi-task reinforcement learning approach on FLUX Fill [dev], we introduce and open-source FLUX Fill [dev][OneReward], a generalized image editing model that outperforms the original model on both inpainting and outpainting tasks, 
    serving as a powerful new baseline for future research in unified mask-guided image generation.
\end{enumerate}

\begin{figure}[p]
    \centering
    \includegraphics[width=\textwidth]{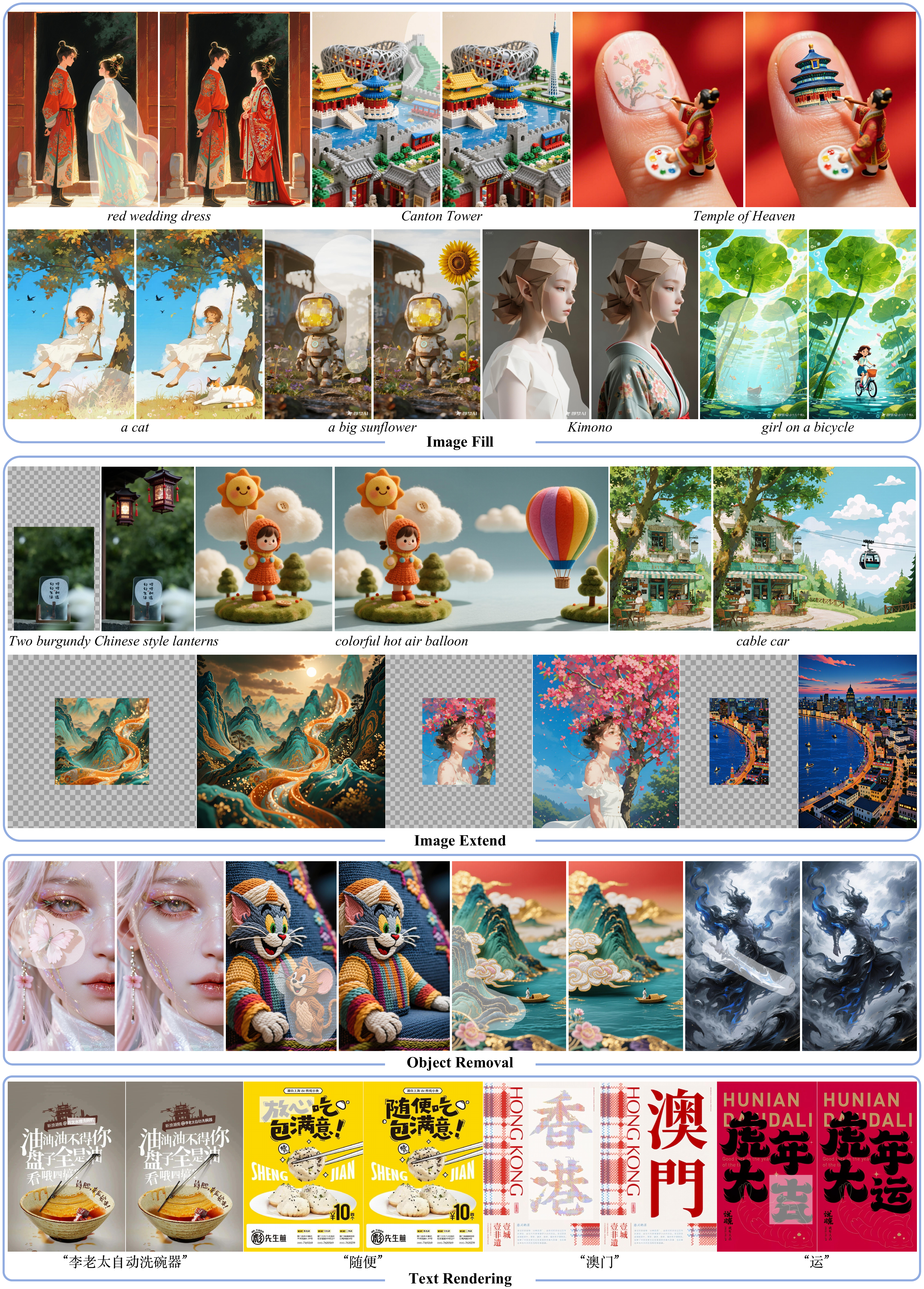}
    \caption{Visual showcase of Seedream 3.0 Fill results across four scenario: image fill, image extend, object removal and text rendering. Each column presents a representative example with corresponding prompts and outputs, demonstrating the model’s unified capability across diverse generation objectives.}
    \label{fig:examples}
\end{figure} 

%% file: relate_work.tex
\section{Related Work}
\textbf{Mask-guided image generation}: Image inpainting and outpainting
focus on generating coherent and seamless content for missing or external regions of an image.
With the advent of deep learning, methods based on Generative Adversarial Networks (GAN)(\cite{goodfellow2020generative}) became dominant. Notably, Large Mask Inpainting(LaMa)(\cite{suvorov2022resolution}) introduced Fast Fourier Convolutions, significantly improving the ability to handle large, complex masks while preserving global structural consistency, a common failure point for earlier CNN-based methods.
More recently, diffusion models(\cite{ho2020denoising, song2020denoising, rombach2022high,song2020score}) have become the state-of-the-art due to their superior generative quality. RePaint (\cite{lugmayr2022repaint}) was an early method that applied a pre-trained unconditional diffusion model to inpainting by repeatedly sampling the unknown region and blending it with the known context, although its iterative nature can be computationally intensive. Subsequent models, such as the native inpainting variant of Stable Diffusion(\cite{rombach2022high,podell2023sdxl}), adopted a more efficient approach by concatenating the latent representations of the mask and the original image as input to its origin text-to-image model. This paradigm established a strong foundation for high-fidelity, text-guided editing.
Follow-up works, such as MagicBrush(\cite{zhang2023magicbrush}) and Inst-Inpaint (\cite{yildirim2023inst}), introduced more refined instruction-based datasets to improve the accuracy of image editing. ByteEdit(\cite{ren2024byteedit}) explored the use of feedback learning to boost performance in these tasks but separate SFT and RL procedures are applied in different sub-task. Recently, FLUX Fill(\cite{flux2024}) has emerged as a powerful open source baseline demonstrating strong performance in both inpainting and outpainting. However, these models are often specialized or lack robust generalization across multiple, distinct editing modalities. Our unified edit model builds directly upon these foundations, but addresses their limitations by leveraging a novel multi-task RLHF framework, unify inpaint, outpainting, object removal, and text rendering within a single, proficient model.

\textbf{RLHF for diffusion model}: Aligning generative models with human preferences has emerged as a rapidly advancing research area, aiming to enhance the aesthetic quality, instruction alignment, and overall user expectation of generated visual content. The success of RLHF critically depends on the quality of the reward model. ReFL (\cite{xu2023imagereward}) makes an important step toward general-purpose reward modeling by training on a large-scale dataset of expert comparisons. It further proposes an algorithm to directly fine-tune diffusion models by treating reward scores as human preference losses and backpropagating them to randomly selected later steps in the denoising process.
Subsequent research, such as VisionReward (\cite{xu2024visionreward}), has explored more fine-grained, multi-dimensional reward modeling by decomposing human preferences into interpretable axes such as fidelity, composition, safety and text alignment. However, its reliance on logistic regression to weight these dimensions introduces additional complexity, limiting its applicability in fully end-to-end training pipelines and reducing generalizability to broader scenarios.
Adapting RLHF algorithms from the large language model (LLM) domain to diffusion models presents a set of unique challenges. Direct Preference Optimization (DPO) (\cite{rafailov2023direct}) was proposed as a simpler and stable alternative to the full RL pipeline. Instead of relying on explicit reward model, DPO optimizes the policy model by directly maximizing the difference in log-probability ratios between preferred and dispreferred responses. This method was effectively extended to the visual domain with Diffusion-DPO (\cite{wallace2024diffusion}), which reformulates the objective in terms of diffusion model likelihood, enabling direct and stable preference alignment. Denoising Diffusion Policy Optimization (DDPO) (\cite{su2024ddpo}) was a pioneering work that successfully applied policy gradient methods to diffusion models by casting the denoising process as a multi-step decision-making problem. Further algorithmic advances include Group Relative Policy Optimization (GRPO) (\cite{shao2024deepseekmath}), which has demonstrated strong performance in aligning both diffusion and flow matching models, as evidenced by its applications in FlowGRPO (\cite{liu2025flow}) and DanceGRPO (\cite{xue2025dancegrpo}).

OneReward synthesizes recent advances in alignment strategies into a unified framework. Our work pushes this frontier further by leveraging only one VLM as the generative reward model to produce task-aware feedback for our multi-task reinforcement learning. It addresses a key limitation of traditional algorithms such as DPO, which struggle to distinguish the winner from the loser when preferences vary across different evaluation dimensions. With OneReward, we develop a SOTA image editing model that jointly learns multiple sub-tasks within a unified reinforcement learning framework.

%% file: preliminaries.tex
\section{Preliminaries}
\subsection{Flow Matching}

Flow Matching(\cite{lipman2022flow}) represents a new class of generative models that offers a more efficient and stable training paradigm compared to traditional diffusion models. Instead of learning the score function of a data distribution, flow matching models learn a velocity vector field that transports a simple prior distribution (e.g., normal distribution) to a complex data distribution through a continuous normalizing flow (CNF).
A CNF is defined by an ordinary differential equation (ODE) that describes the trajectory of a sample $\mathbf{x}$ over a continuous time variable $t\in [0, 1]$:
$\frac{d\mathbf{x}_t}{dt} = v_t(\mathbf{x}_t)$.
Here, $x_0$ is sampled from the prior distribution $p_0$, and $x_1$ follows the target data distribution $p_1$. The function $v_t(x_t)$ is the time-dependent vector field. Flow matching aims to train a neural network $v_{\theta}(x, t, c)$ (where $c$ represents conditioning information such as text or binary mask) to approximate a target vector field $u_t(x|c)$. The conditional flow matching loss is formulated as a simple regression objective:
\begin{center}
$\mathcal{L}_{\text{FM}}(\theta) = \mathbb{E}_{t, p_t(\mathbf{x}|\mathbf{c}), \mathbf{c}} \left[ \left\| v_{\theta}(\mathbf{x}, t, \mathbf{c}) - u_t(\mathbf{x}|\mathbf{c}) \right\|^2 \right]$
\end{center}

Rectified Flow(\cite{liu2022flow}) is a powerful special case of flow matching that linearizes the transport paths to maximize sampling efficiency. It simplifies the objective by defining the target vector field as the constant direction between a data sample $x_1$ and its corresponding noise sample $x_0$, such that $u_t(x|c)=x_1-x_0$.
This formulation avoids the complex score-matching objective of diffusion models, often leading to faster convergence and more efficient generation. Our model, is trained upon this efficient trainging methodology.

\subsection{Reinforcement Learning from Human Feedback }
RLHF is a powerful technique for aligning generative models with complex, hard-to-specify human preferences. 
Human preference data is collected, typically in the form of comparisons. For a given input $c$, annotators are shown two outputs, winner $x^w$ and loser $x^l$, and asked to choose which one they prefer. This pairwise preference dataset is used to train a reward model $r_{\phi}(c, x)$ that predicts a scalar score reflecting human preference. The reward model is often trained using a binary cross-entropy loss based on the Bradley-Terry(\cite{bradley1952rank}) model, which states that the probability of preferring $x^w$ over $x^l$ is:
\begin{center}

$P(x^w \succ x^l | c) = \sigma(r_{\phi}(x^w, c) - r_{\phi}(x^l, c))$
  \end{center}

  where $\sigma$ is the sigmoid function.

ReFL(\cite{xu2023imagereward}) proposes a direct gradient-based fine-tuning method using scalar rewards from a pre-trained reward model. Unlike language models, diffusion models lack a tractable likelihood for complete samples, making traditional RL-based optimization less straightforward to apply. ReFL circumvents this by leveraging the insight that partially denoised samples at latter timesteps already exhibit distinguishable reward scores.
During training, ReFL randomly selects a late denoising step $t$, predicts the corresponding image $x^{\prime}_0$, and computes a scalar reward $r(c, x^{\prime}_0)$ from the reward model. The gradient is then backpropagated through the denoising step using a truncated reward:
$$
\mathcal{L}_{\text{reward}} = \mathrm{ReLU}\left( r(c, \hat{x}_0) \right)
$$
By applying feedback directly on generation trajectories, ReFL enables preference-driven fine-tuning of diffusion models in a scalable and model-agnostic manner.

%% file: reward.tex
\section{METHODOLOGY}
\subsection{Overview}
\begin{figure}[t]  
    \centering  
    \includegraphics[width=1.0\textwidth]{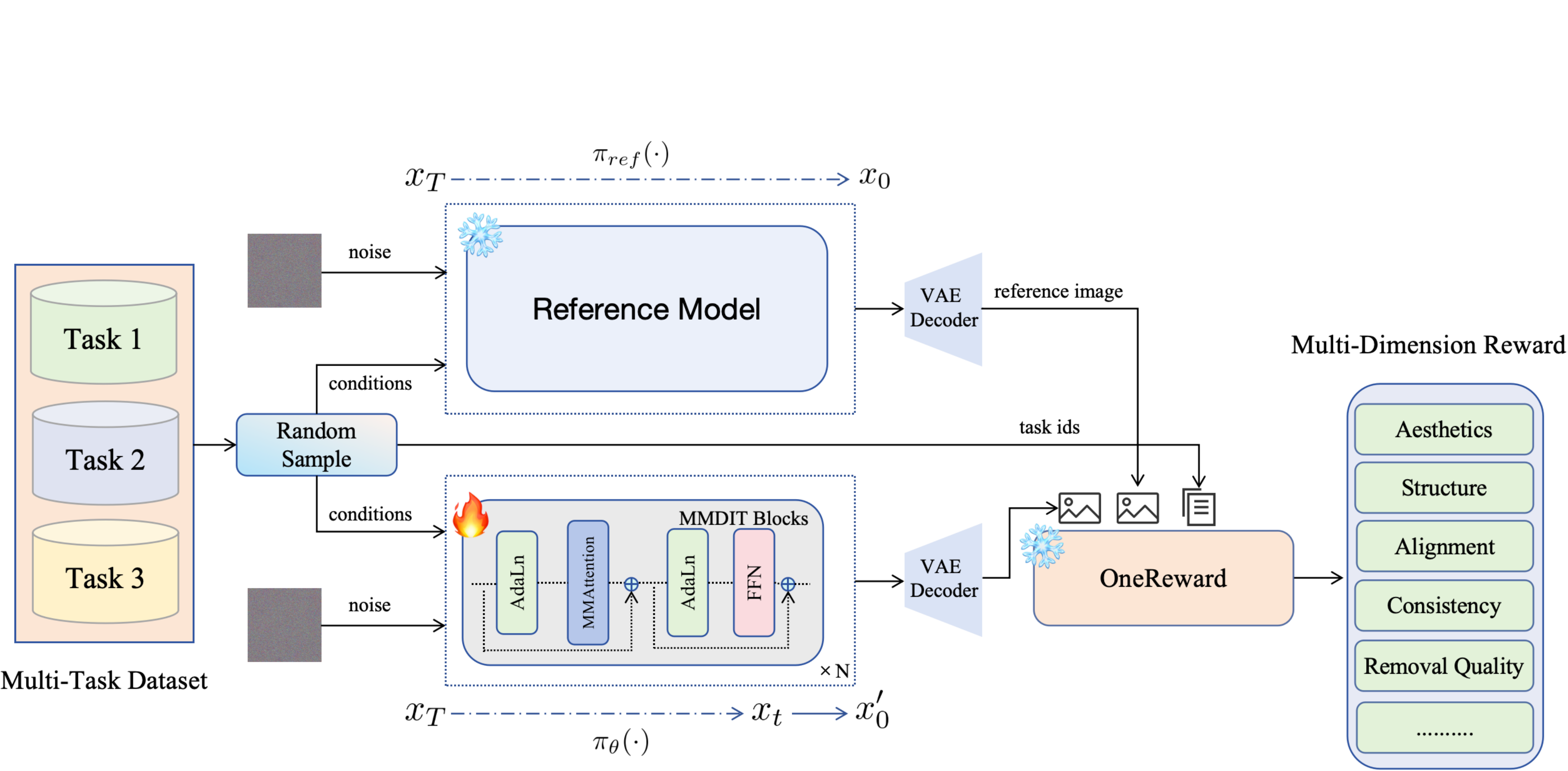}  
    \caption{Overall pipeline of our unified RL procedure. We first random sample image and conditions from different task with a certain probability. Start with same condition and different init noise, the reference image is fully denoised using the reference model, denoted as $\pi_{ref}(\cdot)$. While the evaluation image is partially denoised with randomly selected step and directly predict $x_0^{\prime}$ based on the policy model, denoted as $\pi_{\theta}(\cdot)$. The reward model guides learning by encouraging the policy model to achieve superior performance to the reference model across all evaluation dimensions and tasks.}  
    \label{fig:pipeline}  
\end{figure}
As show in Fig. ~\ref{fig:pipeline}, we introduce a unified framework for multi-task learning, leveraging reward model $r$ trained with human preferences data to fine-tunes the policy model $\pi_{\theta}$ across a range of downstream tasks. We assume that the complete dataset $\mathcal{D} = \{D_k\}_{k=1}^K$ can be partitioned into $K$ subsets. 
For each task, $\mathcal{D}_k= \{(x_i, c_i, s_k, \mathcal{E}_k)\}_{i=1}^N$ share a common format for the image $x_i\in\mathbb{R}^{H\times W\times 3}$ and corresponding input condition $c_i$. The subset $\mathcal{D}_k$ is characterized by a unique identifier $s_k$ and a set of evaluation metrics $\mathcal{E}_k$. Note that $\mathcal{D}_k$ is associated with a distinct set of evaluation criteria, i.e., $\mathcal{E}_k$ may vary significantly across tasks.
Given current task ids $s_k$ and a certain evaluation metric $e\in \mathcal{E}_k$,
the reward model $r$ will be used to calculate the probability how the evaluation image $x_{\theta}$ is better than reference image $x_{ref}$, which are generated by policy model $\pi_{\theta}$ and reference model $\pi_{ref}$, respectively.
In the process of our RL pipeline,
The objective of our training scheme is to increase the probability of generated reward-aligned tokens, as determined by the reward model $r$.
OneReward enables the model to efficiently learn a versatile, multi-task generation policy that satisfies multidimensional human preferences within a single, unified process.

This section is structured as follows: we first detail our data construction pipeline in Section~\ref{sec:data}. Next, in Section~\ref{sec:onereward}, we describe the training procedure for our reward model, termed OneReward. Finally, we present the multi-task, multi-criteria reinforcement learning strategy in Section~\ref{sec:refl}.

\subsection{Human Preference Data Collection}\label{sec:data}
To support the development and evaluation of our unified image editing framework, we constructed a large-scale, high-quality human preference dataset  for multi-task image generation. This dataset spans four major editing tasks: image fill, image extend, object removal, and text rendering, each presenting unique challenges and requiring distinct evaluation perspectives.

In image fill and image extend, the model is required to synthesize plausible content in user-specified regions, guided by natural language prompts. These prompts typically describe the content to be generated, along with contextual details to ensure stylistic and semantic consistency with the surrounding image. In contrast, the object removal task centers on the elimination of specified  elements from the input image. Its objective is not to insert new content, but rather to achieve visually seamless completion of the masked area. Since removal does not involve user-defined content, there is no need to provide a unique prompt as condition for each sample. To ensure input format consistency across all tasks, we apply a fixed, generic prompt (e.g., “remove the specified object”) to all object removal instances.
Each sample is structured as a triplet $(I_{src},M,P)$, where 
$I_{src}\in \mathbb{R}^{H\times W\times 3}$ is the source image, 
$M\in\mathbb{R}^{H\times W} $ is the binary mask indicating the region to be edited, and $P$ is the corresponding text prompt. This unified input representation enables consistent processing across different tasks within our framework.
In the data collection pipeline, we employ a pre-trained diffusion model as the base generator and produce a set of candidate images for each sample by varying key inference parameters, thereby promoting diversity in output quality. This parameter randomization introduces controlled variability across the candidates, enabling the learning of nuanced human preferences. Specifically, we randomly sample the following parameter in the diffusion inference pipeline:
\begin{itemize}
    \item Inference Steps: Uniformly sampled from different denoise steps, controlling the quality of generation.
    \item Negative prompt: Including optional descriptors like ``nsfw", ``blurry" or ``low quality", which guide the generator away from undesired content.
    \item Classifier-Free Guidance Scale: Sampled from a range of values, controlling the trade-off between prompt adherence and creativity.
    \item Initial Noise Strength: Sampled from a predefined range to simulate varying levels of denoising starting point.
\end{itemize}

To meet the needs of multi-tasking and multi-dimensional evaluation, we design a task-specific annotation protocol: 
\begin{itemize}
    \item Structure: Measures whether the generated content maintains spatial and geometric coherence with the original image layout, such as object contours, perspective, and scene topology.
    \item Consistency: Evaluate how smoothly the edited region integrates with its surroundings in terms of color, texture, and illumination, ensuring visual seamlessness.
    \item Text Alignment: Assesses how accurately the generated image content aligns with the semantics of the input prompt, especially in terms of object identity, attributes, and positioning.
    \item Aesthetic: Reflects the overall visual appeal of the result, including composition, realism, and fidelity to high-quality image standards.
    \item Removal Quality: Judging the effectiveness of object removal, focusing on whether the target is completely eliminated and whether the area is cleanly filled without visible artifacts or unintended content.
\end{itemize}

\begin{figure}[!t]
  \centering
  \includegraphics[width=\textwidth]{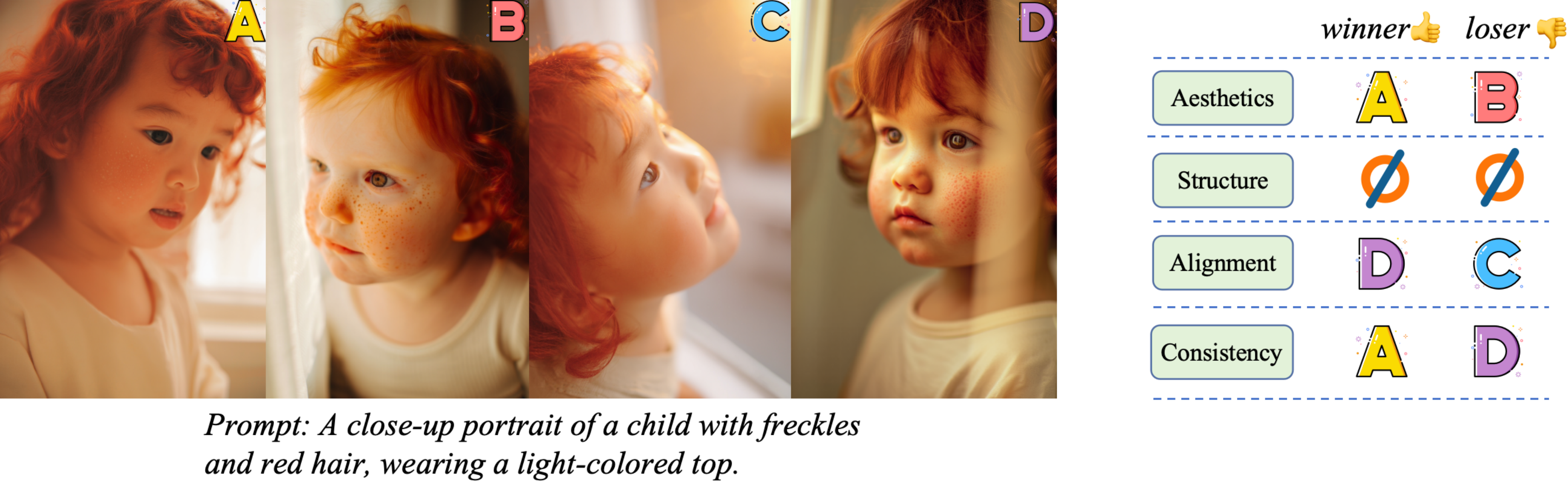}
  \caption{Illustration of the pairwise annotation process. Given multiple
  candidate outputs for the same prompt and binary mask, annotators identify the
  best and worst samples under each evaluation dimension to form a winner/loser
  pair. If the differences between candidates are negligible, the dimension is
  discarded (denoted by $\emptyset$), ensuring that only
  informative comparisons are retained. To clarify, this showcase uses an all-one mask, meaning the entire image region is generated.}
  \label{fig:data_label}
\end{figure}

For image fill and image extend, each set of candidate images is evaluated along the first four dimensions. And for object removal, where no user text is present, we reduce the evaluation to the final single criterion.
As shown in Fig.\ref{fig:data_label}, annotators are presented with the input triplet $(I_{src},M,P)$, and $N$ generated candidate images. For each evaluation dimension, they are instructed to perform a Best-of-$N$ and Worst-of-$N$ selection, identifying the most and least preferred outputs. This process yields a winner/loser pair for each dimension of one sub-task, enabling fine-grained and dimension-specific preference supervision.
Since judgments are made independently across multiple dimensions, the same candidate may be selected as a winner in one dimension and a loser in another. For instance, a particular output may exhibit stronger semantic alignment with the prompt than another candidate, while simultaneously being less aesthetically appealing. The design retains disagreements across evaluation dimensions, instead of forcing them into a fixed but potentially misleading label.

By collecting preference pairs at the metric level, rather than through holistic or averaged scoring, our annotation strategy overcomes a major limitation of prior work by enabling the disentanglement of conflicting judgments across distinct evaluation dimensions. This comparative and multidimensional evaluation scheme promote high-quality and robust human perference data, which serves as a strong foundation for training reward models in multi-objective image generation tasks.

\subsection{Multi-Dimensional Pairwise Reward Model Training}\label{sec:onereward}
\begin{figure}[htbp]
  \centering
  \includegraphics[width=\textwidth]{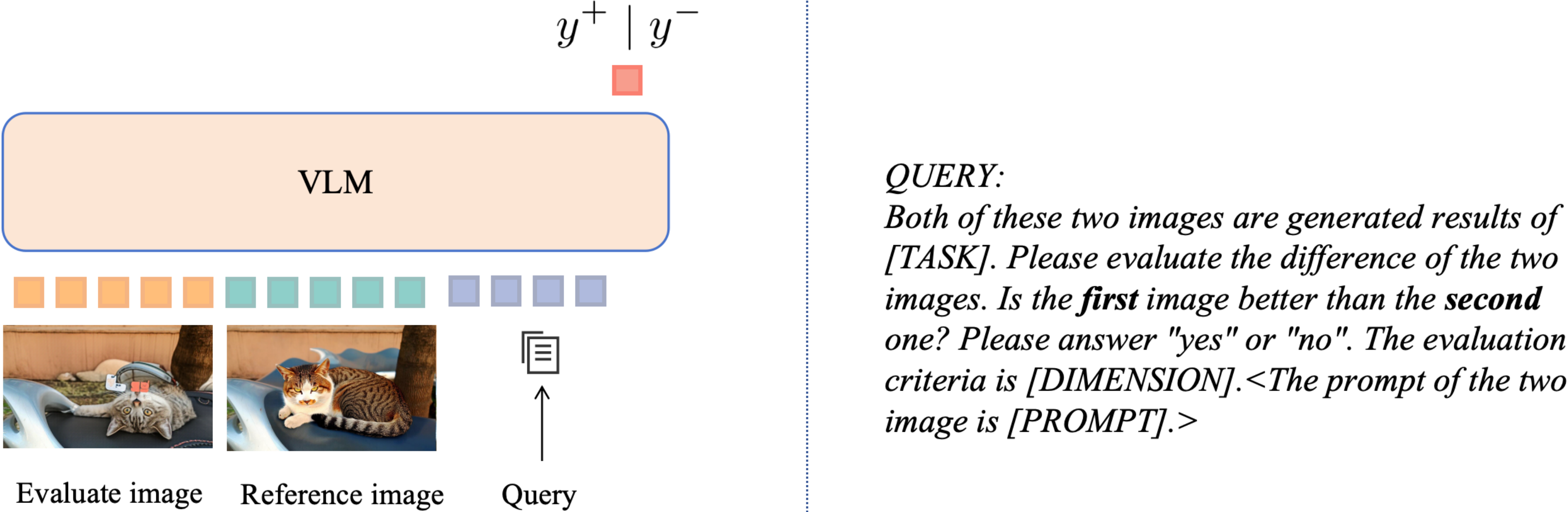}
  \caption{The detail of our one reward model. We utilize VLM to judge whether the first image is better than the second one. In the process of  reward feedback learning, the probability of $y^+$ token is treated as the reward to the diffusion models. We simplely add the edit task and the evaluation dimensions to the user query, achieving the goal of training for different task and dimensions. The content of angle brackets is optional, only add when the evaluation dimension is  Text Alignment.}
  \label{fig:rm_train}
\end{figure}
Designing a multi-task learning framework for image editing presents significant challenges, particularly in constructing a reward mechanism that is both scalable across heterogeneous tasks and robust against reward hacking. A naive solution, training separate reward models for each evaluation dimension within each task, is computationally expensive and difficult to tune, requiring considerable resources for both training and optimization. Conventional scalar-based reward models are particularly inadequate for mask-guided generation tasks, as their assessments are easily dominated by unchanged background regions. As a result, they fail to capture the true quality of the edited content within the masked area, producing reward signals that are misaligned with the actual effectiveness of the intended edits.

To address these limitations, we propose OneReward, a unified reward model designed to assess outputs across multiple image editing tasks and evaluation dimensions with only one reward model. OneReward is built upon the perceptual capabilities of a pre-trained VLM, which serves as the backbone for all-dimension reward prediction. Rather than introducing task-specific output heads, we guide the model through a textual evaluation query $q$ , which encodes both the task identifier $s_k$ and the evaluation dimension  $e \in \mathcal{E}_k$. This formulation enables dynamic conditioning of the VLM on the specific aspect of quality to be evaluated. Formally, the evaluation query is constructed as:
\begin{equation}
q = \Phi(s_k, e, P)
\end{equation}
where  $\Phi$ is a instruction template as shown in Fig.~\ref{fig:rm_train}, $P$ is the original input prompt, used only when evaluating prompt–image alignment. For all other dimensions, such as aesthetics or structural, the query remains prompt-free, ensuring that judgments reflect intrinsic visual quality without semantic bias.

To further enhance robustness and address the shortcomings of absolute scoring, OneReward employs a comparative evaluation scheme. 
It receives a pair of generated images and is asked to identify the preferred one, based on relative quality.
As illustrated in Fig.~\ref{fig:rm_train}, the reward model takes a pair of images ($x^w, x^l$) along with an evaluation query $q$ as input, and outputs a binary classification that aligns with the human preference for that specific query. 
It is defined as:
\begin{equation}
y = r(x^w, x^l, q)
\end{equation}

Our training methodology derives directly from the Best-of-N and Worst-of-N annotation protocol in Sec.~\ref{sec:data}. 
Specifically, for each set of four candidate images, annotators independently select the best image $x^{w}$ and the worst one $x^{l}$ for each evaluation dimension, forming pairs of human preference for training. 
This process allows each preference pair to be labeled with one or more evaluation dimensions, resulting in a multi-label learning scenario. 
For instance, in Fig.~\ref{fig:data_label}, the fourth image excels in text alignment, but underperforms in consistency. As a result, it is labeled as the winner for text alignment and the loser for consistency.
For each training sample, we construct the corresponding evaluation queries $q$ for all the annotated dimensions and compute the standard cross-entropy loss.
The loss function is defined as follows:

\begin{equation}
\mathcal{L}_{}(\phi) =-\frac{1}{2}\mathbb{E}_{\left(x^w, x^l, q\right) \in \mathcal{D}}\left[\log\mathbb{P}_\phi \left(y^+ \mid x^w,\; x^l,\; q\right)+\log\mathbb{P}_\phi\left(y^- \mid x^l,\; x^w,\; q\right)\right]
\end{equation}
where $y^+$ denotes affirmative token (e.g. ``Yes''), $y^-$ denotes negative token (e.g. ``No''), and $\mathbb{P}_\phi$ represents the probability assigned to the corresponding token by the reward model. $x^w$ and $x^l$ refer to the winner and loser pair under a specific evaluation criterion.
Note that this formulation is non-commutative, as the query is designed to expect the response $y^+$ when the image in the first position is of higher quality. Otherwise, the expected response is $y^-$.
This training strategy enables the efficient utilization of our annotations, empowering the reward model with the capacity to perform nuanced, multidimensional evaluations across a range of tasks.

Table~\ref{tab:reward_acc} reports the accuracy of our reward model in different sub-task and evaluation dimensions. The results show that text alignment achieves the highest accuracy, exceeding 80\% in both image fill and image extend. Other dimensions such as consistency, structure, and aesthetics yield moderate but stable accuracies, mostly in the low–mid 70\% range. For the object removal task, the model reaches 84.93\% in removal quality, indicating that it is particularly effective at capturing human preferences in this setting. Since the underlying VLM is pre-trained with multimodal supervision, it is naturally more sensitive to text–image correspondence, making text alignment easier to discriminate, whereas dimensions related to intrinsic visual quality remain more challenging. Overall, these results suggest that the model generalizes well across tasks and metrics.

\begin{table}[t!]
\centering
\resizebox{0.98\textwidth}{!}{%
\begin{tabular}{l|*{5}{c}}
\toprule
\textbf{Accuracy(\%)}&   \textbf{Text Alignment} &\textbf{Consistency} & \textbf{Structure} & \textbf{Aesthetics}&\textbf{Removal Quality}\\
\midrule
Image Fill   & 83.77 & 74.23 & 74.63 & 72.10 & -- \\
\midrule
Image Extend & 80.36 & 72.50 & 74.29 & 71.10 & --\\
\midrule
Object Removal &--&--& --&--& 84.93\\
\bottomrule
\end{tabular}%
}
\caption{Accuracy of the reward model across multiple editing tasks and evaluation dimensions. Each entry denotes the model’s accuracy in distinguishing winners from losers on the test set for a given dimension (e.g., Text Alignment, Consistency, Structure, Aesthetics, Removal Quality). Higher values indicate more reliable preference discrimination by the reward model.}
\label{tab:reward_acc}
\end{table}

\subsection{Multi-Task Reward Feedback Learning}\label{sec:refl}
Based on OneReward, we propose a novel framework that systematically aligns a pre-trained diffusion model with complex, multi-dimensional human preferences across a suite of tasks.
As shown in Fig. ~\ref{fig:pipeline}, the whole train pipeline is composed of three main components: a frozen reference model $\pi_{ref}$, a trainable policy model $\pi_{\theta}$, and the trained OneReward model $r$. In each iteration, we first sample data from the sub-dataset $\mathcal{D}_k$ within the full dataset $\{\mathcal{D}_k\}_{k=1}^K$, according to a predefined sampling probability $p_k$. To promote learning in more difficult tasks, we assign higher sampling probabilities to the task that are estimated to be more difficult based on prior knowledge.
The reference model $\pi_{ref}$ is initialized by the parameters of the pre-trained diffusion model. It generates a baseline image $x_{ref}$  via a full denoising trajectory from gaussian noise $x_T$  to the final output $x_0$, representing the model’s initial generative capacity. Inspired by ReFL (\cite{xu2023imagereward}), the evaluate image $x_{\theta}$ start from the same condition $c$ but with a different noise vector, performing partial denoising up to a randomly selected intermediate timestep $t$ before directly predicting the final latent $x_0^{\prime}$. 
The vae decoder, denote as $\mathrm{vae\_decode}(\cdot)$, is used to reconstructs pixel-space images from these latents and gradients are backpropagated only through this final single-step prediction, making the training process efficient.
Both the prediction of the evaluate $x_{\theta}$ and the reference image $x_{ref}$, along with an evaluation query $q$, are passed into the reward model. The reward model evaluates whether $x_{\theta}$ is preferred over $x_{ref}$ under the given task and evaluation dimension, and returns a binary output token $y^+$ and $y^-$:
\begin{figure}[!t]
    \centering
    \includegraphics[width=1.0\textwidth]{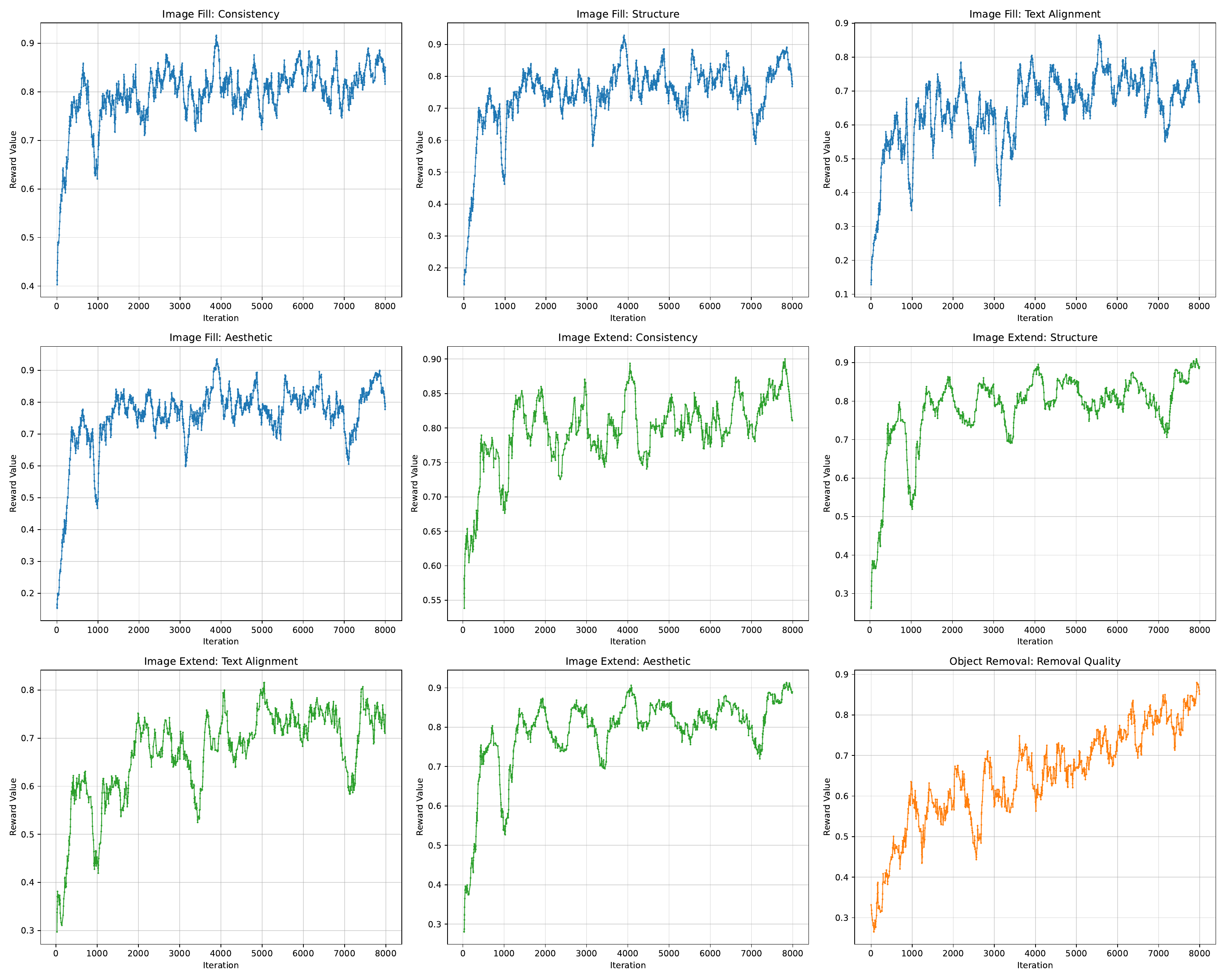}
    \caption{We visualize the reward curves of Consistency, Structure, Text Alignment, Aesthetics for image fill(blue) and image extend(green), Removal Quality for object removal(orange).}
    \label{fig:rl_reward}
\end{figure}

\begin{equation}
\begin{aligned}
x_{\theta}     &= \mathrm{vae\_decode}(x_0^{\prime}) \\
x_{\mathrm{ref}} &= \mathrm{vae\_decode}(x_0) \\
y              &= r(x_{\theta}, x_{\mathrm{ref}}, q)
\end{aligned}
\label{eq:rm_cal}
\end{equation}

As $x_{\theta}$ and $x_{ref}$ are generated from the same condition $c$ and different model parameters, we simplify the rollout procedure as $\pi_{\theta}(\cdot)$ and $\pi_{ref}(\cdot)$, respectively.
In the process of RL, The probability to response $y^+$ is treated as the reward signal. And the objective of the policy model is to maximize this expected reward, the loss function is defined as follows.
\begin{equation}
\mathcal{J}_{}(\theta) = \max\left(0, \lambda - \mathbb{P}_\phi\left(y^+ \mid \pi_\theta(c),\; \pi_{\text{ref}}(c),\; q\right)\right)
\label{equ.rl_loss}
\end{equation}
\noindent\text{where } $y^+$ denotes the affirmative token defined in Sec.~\ref{sec:onereward}  and $\mathbb{P}_\phi$ is the probability predicted by the reward model parameterized by $\phi$, $\lambda$ is the predefined reward upper bound to avoid hacking, the query $q$ contains the information of task ids and its task-specific evaluation dimensions, as shown in Fig.~\ref{fig:rm_train}.
Our training objective for the policy model involves simultaneously optimizing for all relevant evaluation metrics of a given task. At each step, dimension-wise rewards are computed and averaged to guide optimization, effectively casting the training process as a form of multi-objective reinforcement learning. This encourages the model to achieve balanced improvements across diverse evaluation criteria, rather than overfitting to any single aspect of generation.A detailed description of the multi-task reinforcement learning process can be found in Alg.~\ref{alg:rl}.

Fig.~\ref{fig:rl_reward} illustrates the reward curves across different evaluation dimensions during training, covering three tasks: image fill (blue), image extend (green), and object removal (orange). Each subplot shows the evolution of reward values over training iterations for specific task–dimension pairs, including Consistency, Structure, Text Alignment, Aesthetics, and Removal Quality. Compared with single-task settings, multi-task training introduces moderate fluctuations in the reward trajectories. Nevertheless, all curves display a clear upward trend, indicating consistent performance gains across dimensions.
These results suggest that while multi-task optimization inherently introduces greater variability due to shared capacity and competing objectives, the model still converges reliably with steady improvements. The effectiveness of our reward-driven training scheme is further validated by the high final reward values achieved across all dimensions.

\makeatletter
\newcommand{\algorithmicoutput}{\textbf{Output:}} 
\newcommand{\OUTPUT}{\item[\algorithmicoutput]}   
\makeatother

\begin{algorithm}[!t]
    \caption{Multi-Task Reinforcement Learning from Human Feedback}
    \label{alg:rl}
    \renewcommand{\algorithmicrequire}{\textbf{Dataset:}}
    \renewcommand{\algorithmicensure}{\textbf{Input:}}
    
    \begin{algorithmic}[1]
        \REQUIRE Multi-Task image-condition datasets
        $\{\mathcal{D}_k\}^K_{k=1}$, with data sample probability distribution $\mathcal{P} = \{p_1, p_2, \dots, p_K\}$, task ids $S = \{s_1, s_2, \dots, s_K\}$ and each evaluation dimension $\{\mathcal{E}_k\}^K_{k=1}$. 
        \ENSURE Reference diffusion model $\pi_{ref}$, policy model $\pi_{\theta}$ with parameters $\theta$, unified reward model $r$ with parameters $\phi$, hyperparameters $[t_1, t_2]$ for the generation of evaluate image.

        \STATE Init reference model $\pi_{ref} \leftarrow \pi_{\theta}$
        \STATE Init ema model $\pi_{ema} \leftarrow \pi_{\theta}$
        \FOR {iteration = 1, \dots, N}

        \STATE Sample condition $c$ from the $k$-th dataset $\mathcal{D}_k $with probability $p_k$
        \STATE Sample init noise $\epsilon_1, \epsilon_2$ from normal distribute $\mathcal{N}(0, 1)$
        \STATE Random sample denoise timesteps $t$ from $[t_1, t_2] $  
        \STATE Generate the reference image $x_{ref}$ with full denoise procedure $\pi_{ref}(\epsilon_1,c)$   
        \STATE Generate the evaluate image $x_{\theta}$ with random denoise steps $\pi_\theta(\epsilon_2,c, t)$
        \FOR {$e\in \mathcal{E}_k$}
            \STATE Generate query $q$ with task id $s_k$ and current evalution dimension $e$ as shown in Fig.\ref{fig:rm_train}
            \STATE Compute RL loss $\mathcal{J}_e(x_\theta,x_{ref},q)$ in Equation\ref{equ.rl_loss} with reward model $r$
        \ENDFOR
        
        \STATE Updata policy model via gradient ascent:$\pi_{\theta}\leftarrow \pi_{\theta}+\frac{1}{|\mathcal{E}_k|}\nabla_{\pi_{\theta}}\sum\limits_{e \in \mathcal{E}_k}\mathcal{J}_e$
        \STATE EMA update $\pi_{ema} \leftarrow \tau\pi_{ema} +(1-\tau)\pi_{\theta}$
        \ENDFOR

        \OUTPUT $\pi_{\theta}, \pi_{ema}$
    \end{algorithmic}
\end{algorithm}

%% file: experiments.tex
\section{Experiments}
\subsection{Implementation Details}
\textbf{Dataset}. Our dataset for multi-task image editing was constructed through a multi-stage pipeline. We began by extracting image embeddings using the CLIP model(\cite{radford2021learning}), followed by K-means clustering to obtain a diverse and representative subset for reinforcement learning. 
Based on the selected subset, we further generated and annotated a large number of human preference pairs, which serve as training data for both the reward model and the image generation model.

\textbf{Experimental Settings}. For our reward model, we fine-tuned the open-source Qwen2.5-VL-7B-Instruct(~\cite{Qwen2.5-VL}) on our preference dataset with batch size 16, learning rate 1e-6. For Seedream 3.0 Fill, We adopt Seeddream 3.0(~\cite{gao2025seedream}) as the text-to-image base model, data were sampled with probabilities of 50$\%$ for image fill, 25$\%$ for image extend, and 25$\%$ for object removal. Training was performed with batch size 8, learning rate 1e-5.

\subsection{Comparisons with State of the arts}

\begin{table}[!t]
\centering
\resizebox{0.98\textwidth}{!}{%
\begin{tabular}{l|l|*{8}{c}}
\toprule
\textbf{Task}& \textbf{Model} & \textbf{Usability Rate(\%)} & \textbf{Text Alignment} & \textbf{Texture Consistency} & \textbf{Style Consistency} & \textbf{Structure} & \textbf{Aesthetics}& \textbf{Text Rendering(\%)} &\textbf{Removal Quality(\%)}\\
\midrule
\multirow{5}{*}{Image Fill} 
&Adobe Photoshop &45.22& 4.09&4.05& 3.83 &2.89& 2.16 &26.69&-- \\
&Ideogram & 50.65 &3.78& 4.25& \textbf{4.13} &3.80& 2.65 &24.81 &--\\
 & Flux Fill [pro] &50.97& 4.12 &4.16& 3.72 &3.46 &2.47 &29.32 &-- \\
& Higgsfield &52.11& 4.43& 4.29& 3.63& 3.54& 2.40 &45.49&-- \\
&\textbf{SeedDream 3.0 Fill} &\textbf{69.04}& \textbf{4.57}& \textbf{4.33}& 3.76& \textbf{4.02}& \textbf{2.91} & \textbf{70.68} &--\\\toprule
\multirow{4}{*}{\shortstack{Image Extend \\ w Prompt}}  
&Adobe Photoshop &44.07& 4.21& 4.04& \textbf{4.08}& 2.79& 2.27 &--&--\\
 & Flux Fill [pro] &44.26& 4.23& 3.77& 3.79& 3.57& 2.36 &--&--\\
 & Ideogram &63.13& \textbf{4.44}& 3.63& 4.02& 3.80& 2.61 &--&--\\
&\textbf{SeedDream 3.0 Fill} &\textbf{64.72}& 4.26& \textbf{4.19}&3.60& \textbf{4.05}& \textbf{2.89} &--&--\\
\midrule
\multirow{5}{*}{\shortstack{Image Extend \\ w/o Prompt}}
&Adobe Photoshop &61.10&--& 3.98& \textbf{4.49}& 2.93& 2.55&--&--\\
 &Midjourney &69.59&--& 4.24& 4.33& 3.47& 2.95&-- &--\\
 & Flux Fill [pro] &70.47&--& 4.10& 4.37& 3.68& 2.84 &--&--\\
 &  Ideogram &73.71&--& 3.99& 4.40& 3.86& 2.92&--&-- \\
 
&\textbf{SeedDream 3.0 Fill} &\textbf{87.54}&--& \textbf{4.36}& 4.02& \textbf{4.19}& \textbf{3.29}&-- &--\\
\midrule
\multirow{4}{*}{Object Removal}&Flux Fill [pro] &15.92&--& 3.97&--& 3.39& --&--&18.71\\
 & Ideogram &70.14&--& \textbf{4.04}&--& 4.07&--&--&72.00\\
&  Adobe Photoshop &73.98&--& 4.03&--& 3.68&--&--&83.16\\
&\textbf{SeedDream 3.0 Fill}  &\textbf{82.22}&--& 3.87&--& \textbf{4.32}& --&--&\textbf{86.33}\\
\bottomrule
\end{tabular}%
}
\caption{Quantitative comparison of our model against SOTA competitors across four editing tasks. Metrics with percentages (e.g., Usability Rate, Text Rendering, Removal Quality) are reported as success rates, while other dimensions (e.g., Text Alignment, Texture Consistency, Style Consistency, Structure, Aesthetics) are Mean Opinion Scores (MOS) rated on a 1–5 scale. Higher scores indicate better performance in all dimensions. Our model demonstrates consistently superior performance across most dimensions compared with existing SOTA models, especially in overall usability rate.}
\label{tab:compare_result}
\end{table}

\begin{figure}[htbp]
    \centering
    \includegraphics[width=0.95\textwidth]{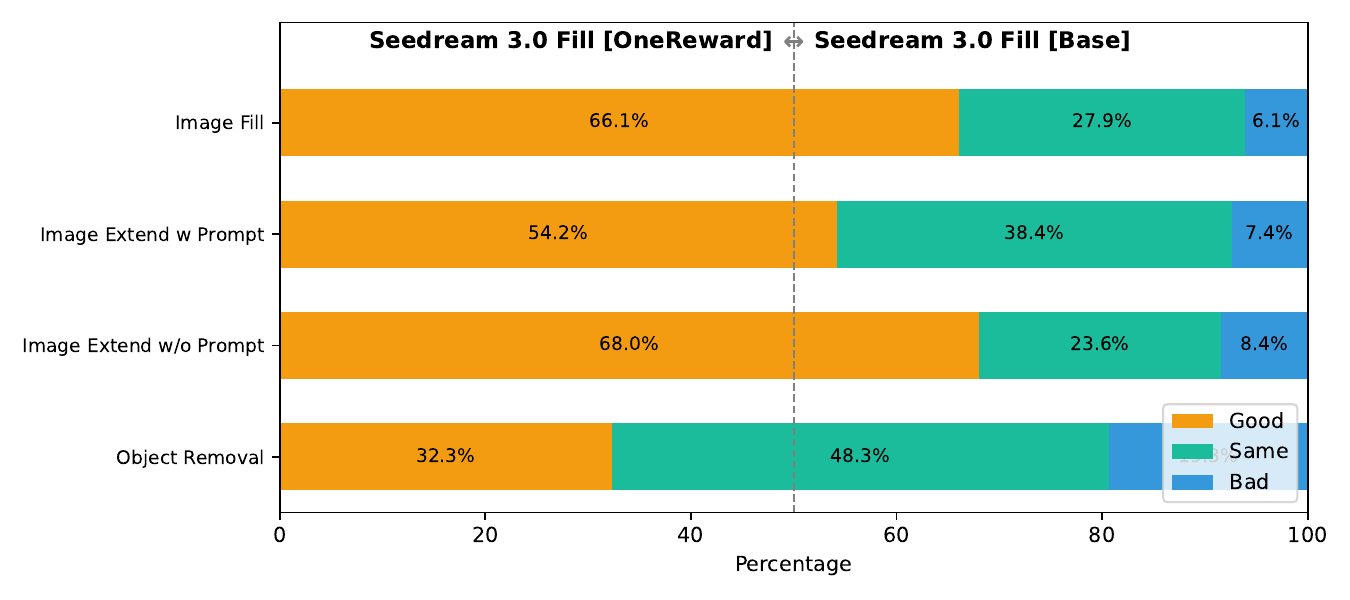}
    \caption{Comparison of performance between Seedream 3.0 [OneReward] and Seedream 3.0 [Base] using Good–Same–Bad (GSB) evaluation. Each group corresponds to a specific image editing task. Bars represent the relative proportions of outputs judged as Good (orange), Same (green), or Bad (blue) across different model pairs. 
    This visualization highlights the distribution of relative preferences, showing where OneReward-enhanced models outperform the base model.}
    \label{fig:gsb}
\end{figure}
\subsubsection{Evaluation Settings}

To comprehensively evaluate the effectiveness of OneReward, we conducted extensive comparisons with SOTA models and APIs, including Ideogram, Higgsfield, Adobe Photoshop, Midjourney, and Flux Fill [Pro]. For evaluation, we constructed a diverse benchmark consisting of 130 images for image fill, 100 for object removal, and 200 for image extend (split evenly between text-guided and text-free). It covers a wide range of scenes, such as portraits, landscapes, pets, and typography, as well as artistic styles such as photorealistic, anime, watercolor, and AI-generated.

\subsubsection{Human Evaluation}
\begin{figure}[!t]
    \centering
    \includegraphics[width=0.9\textwidth]{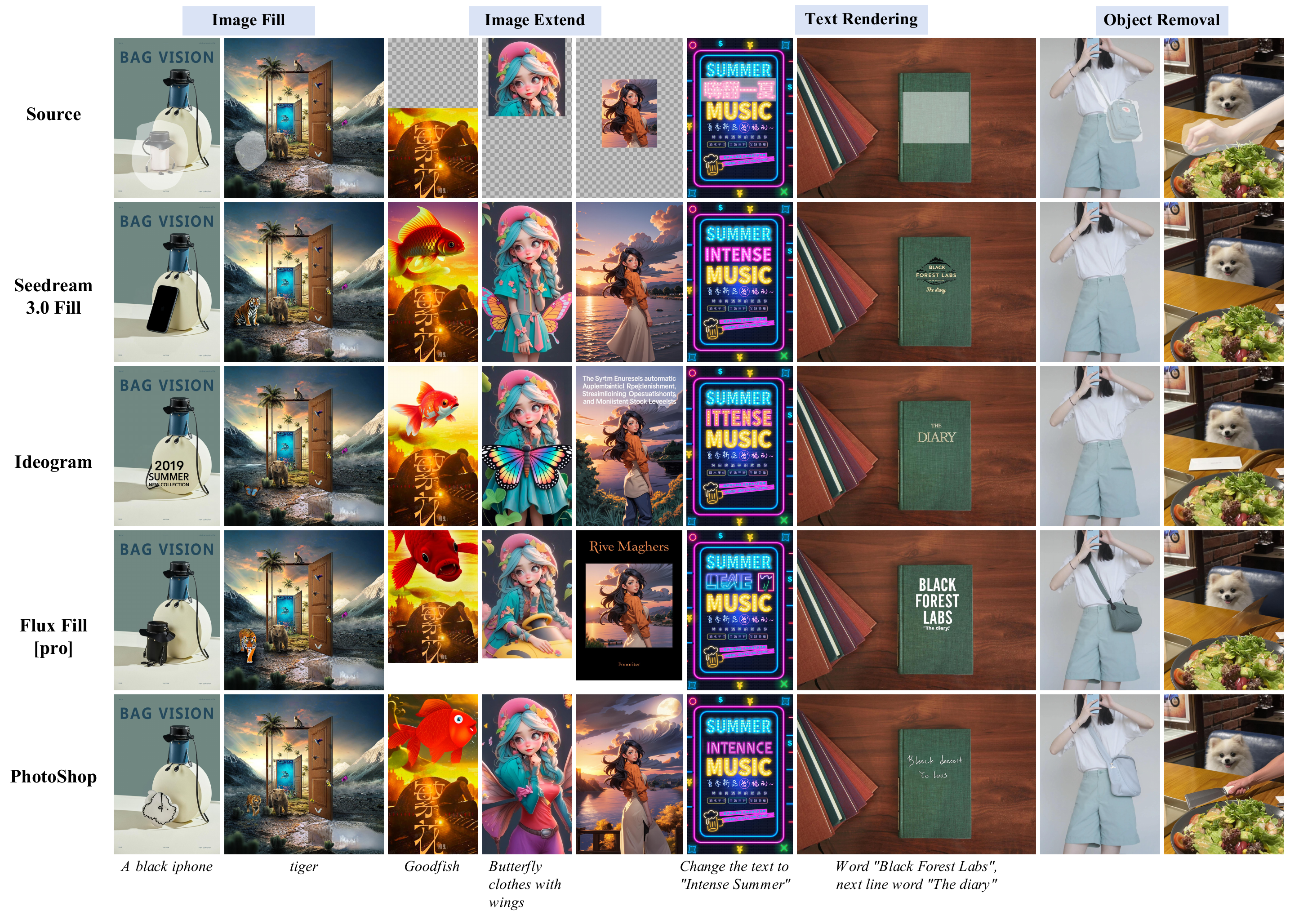}
    \caption{Visual comparison of editing results for Seedream 3.0 Fill and its competitors across different tasks. Rows correspond to different methods, columns show task-specific prompts and outputs. The source images are shown in the first row. The blank row at the bottom indicates that the case is prompt-free.}
    \label{fig:compare}
\end{figure}

To comprehensively evaluate the performance of our model, we conducted a user study involving 40 participants. Each participant rated the generated images across multiple dimensions: overall quality, text alignment, texture consistency, style consistency, structural plausibility, aesthetics, text rendering, and removal quality.
Among these, overall usability, text rendering, and removal quality were treated as binary judgments, where each image was assessed as either acceptable or not. The reported values for these metrics therefore represent success rates expressed as percentages. In contrast, the remaining dimensions were rated on a 1–5 Likert scale, and the scores were averaged to produce Mean Opinion Scores (MOS), where higher values indicate better quality.

Comparative results for the four image editing tasks are presented in Fig.~\ref{fig:radar} and Tab.~\ref{tab:compare_result}. 
SeedDream 3.0 Fill demonstrates the strongest overall performance across all tasks. For image fill, our model achieves a usability rate of 69.04$\%$, outperforming the second-best competitor (52.11$\%$) by 16.93 percentage points. It also obtains the highest scores in most dimensions, including text alignment, texture consistency, structure, aesthetics, and text rendering, with the only exception being style consistency, where Ideogram shows a slight advantage.
On the text-guided image extend task, SeedDream 3.0 Fill performs comparably to Ideogram while clearly surpassing Flux Fill [pro] and Adobe Photoshop in usability. 
And in the text-free setting, SeedDream 3.0 Fill shows pronounced superiority, achieving the highest usability rate (87.54$\%$) and leading across all reported dimensions.
For object removal, our model again delivers the best results, with a usability rate of 82.22$\%$ and a removal quality score of 86.33$\%$, significantly outperforming other SOTA competitors. The high removal quality indicates that our model produces the fewest unwanted objects in this task, behavior that typically conflicts with goals in other generation tasks such as image fill or extend. This demonstrates the effectiveness of our RL strategy in multi-task human preference learning.

To further assess the impact of OneReward, we conduct a Good–Same–Bad (GSB) evaluation comparing SeedDream 3.0 Fill models trained with and without reward guidance. As shown in Fig.~\ref{fig:gsb}, each bar represents the distribution of human preferences across different tasks. Compared to the base model, the OneReward variant receives a higher proportion of “Good” ratings in all tasks.
The GSB results demonstrate that our unified reward model generally shifts model outputs toward preferred generations.

Notably, all of the above results were achieved using a unified generation model, trained with only one reward model shared across different task and evaluation dimension, without any task-specific SFT.

\subsection{Dynamic Reinforcement Learning}
As shown in Alg.~\ref{alg:rl}, our training pipeline typically maintain three models in parallel: a policy model, a reference model, and an EMA variant. This design leads to high memory consumption and increases the engineering complexity of model synchronization. On the other hand, if the reference images are of insufficient quality, the policy model may be trained on overly easy preference pairs, potentially leading to reward hacking and hindering effective learning.

\begin{figure}[!t]
    \centering
    \includegraphics[width=0.9\textwidth]{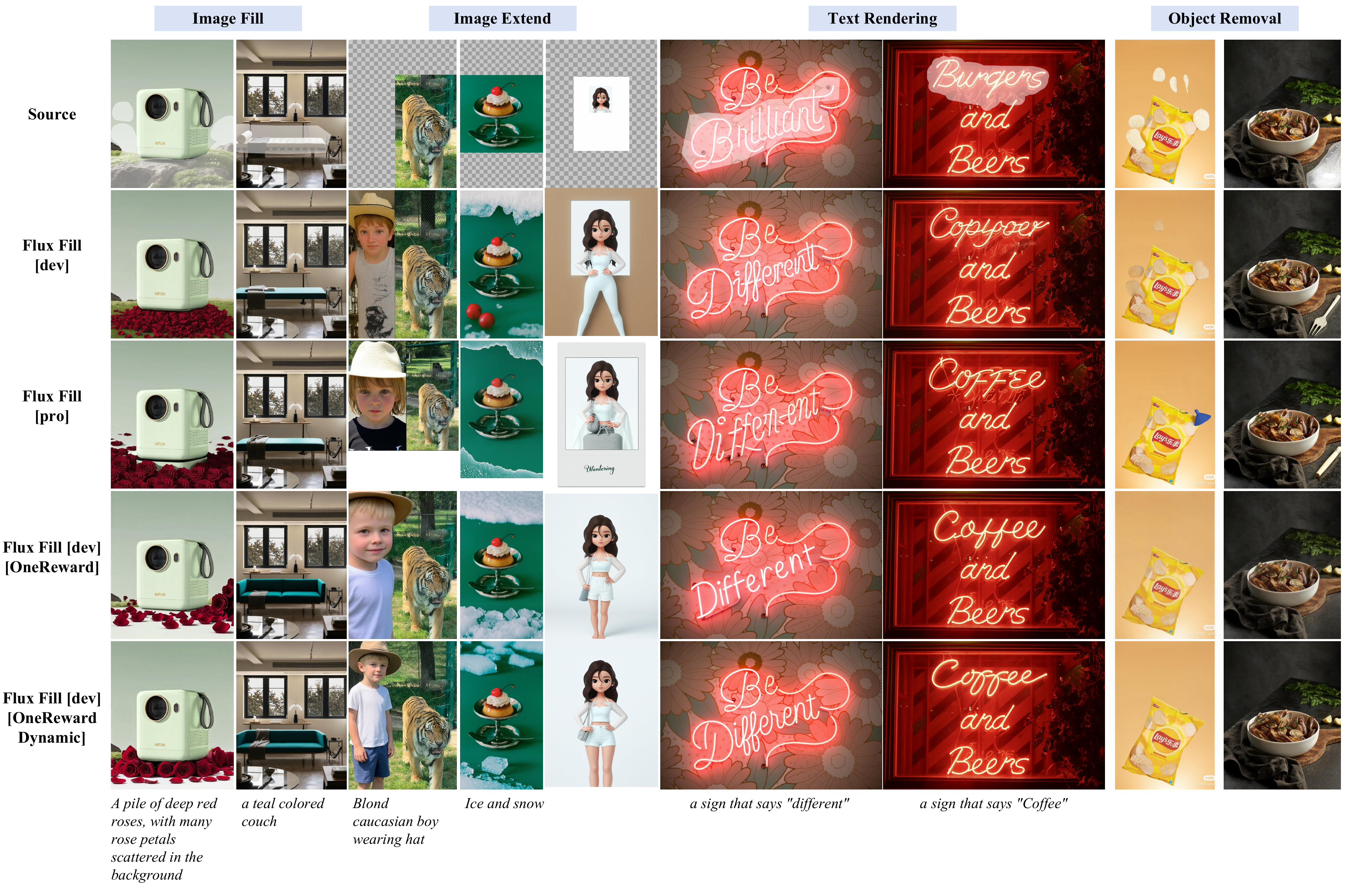}
    \caption{Visual comparison of editing results for Flux Fill and our RL model across different tasks. Rows correspond to different methods, columns show task-specific prompts and outputs. The source images are shown in the first row. The last two rows stands for our RL-enhanced model, trained via Alg.~\ref{alg:rl} and Alg.~\ref{alg:dy_rl}.}
    \label{fig:flux}
\end{figure}

To address these limitations, we propose a dynamic reinforcement learning strategy (Alg.~\ref{alg:dy_rl}), in which the EMA model is directly reused as the reference model. As training progresses, the EMA model gradually improves in generative quality, thereby providing an increasingly strong baseline for policy comparison. This design not only reduces memory overhead by eliminating the need for a separate reference model, but also yields more stable and adaptive reward signals throughout training.
A conceptual visualization of the reward computation process is provided in Figure~\ref{fig:dy_rl}, highlighting the difference between the baseline setup and this dynamic design. In the dynamic variant, the reference model is continuously enhanced during training, ensuring that the policy is always compared against a strong and progressively improving baseline. Beyond its simplicity and efficiency, the dynamic framework demonstrates highly competitive performance, as shown in the last row of Fig.~\ref{fig:flux}.

\subsection{Qualitative Results}
Fig.~\ref{fig:compare} presents visual comparisons in different situations. We compare Seedream 3.0 Fill with Ideogram, Flux Fill [pro], and Adobe Photoshop using the same input conditions.
Based on Flux Fill [dev], we applied both Alg.~\ref{alg:rl} and Alg.~\ref{alg:dy_rl} for RL training. Fig.~\ref{fig:flux} shows a comparison of the resulting models with Flux Fill [dev] and Flux Fill [pro]. It shows that our RL-enhanced model, trained via OneReward, demonstrates superior visual quality compared to the base model and even the closed-source API.
We will release the both trained models as part of our open-source contribution.

%% file: conclusion.tex
\section{Conclusions}
We present OneReward, a unified reward model designed for multi-task, multi-dimensional reinforcement learning of diffusion and flow mathching model. It enables fine-grained supervision across diverse tasks by leveraging VLM as the reward model.
Built on OneReward, our develop Seedream 3.0 Fill, which achieves SOTA performance on image fill, image extend, object removal, and text rendering, outperforming both commercial APIs and open-source models on most dimensions. While style consistency remains a relative weakness, we leave it for next optimization through detailed data annotation and post-training refinement.
To further support the community, we open-source FLUX Fill [dev][OneReward], an enhanced variant of flux model trained with our RL framework, offering stronger generalization and broader task applicability.

%% file: appendix.tex
\section{Appendix}
\begin{figure}[htbp]
    \centering
    \includegraphics[width=0.9\textwidth]{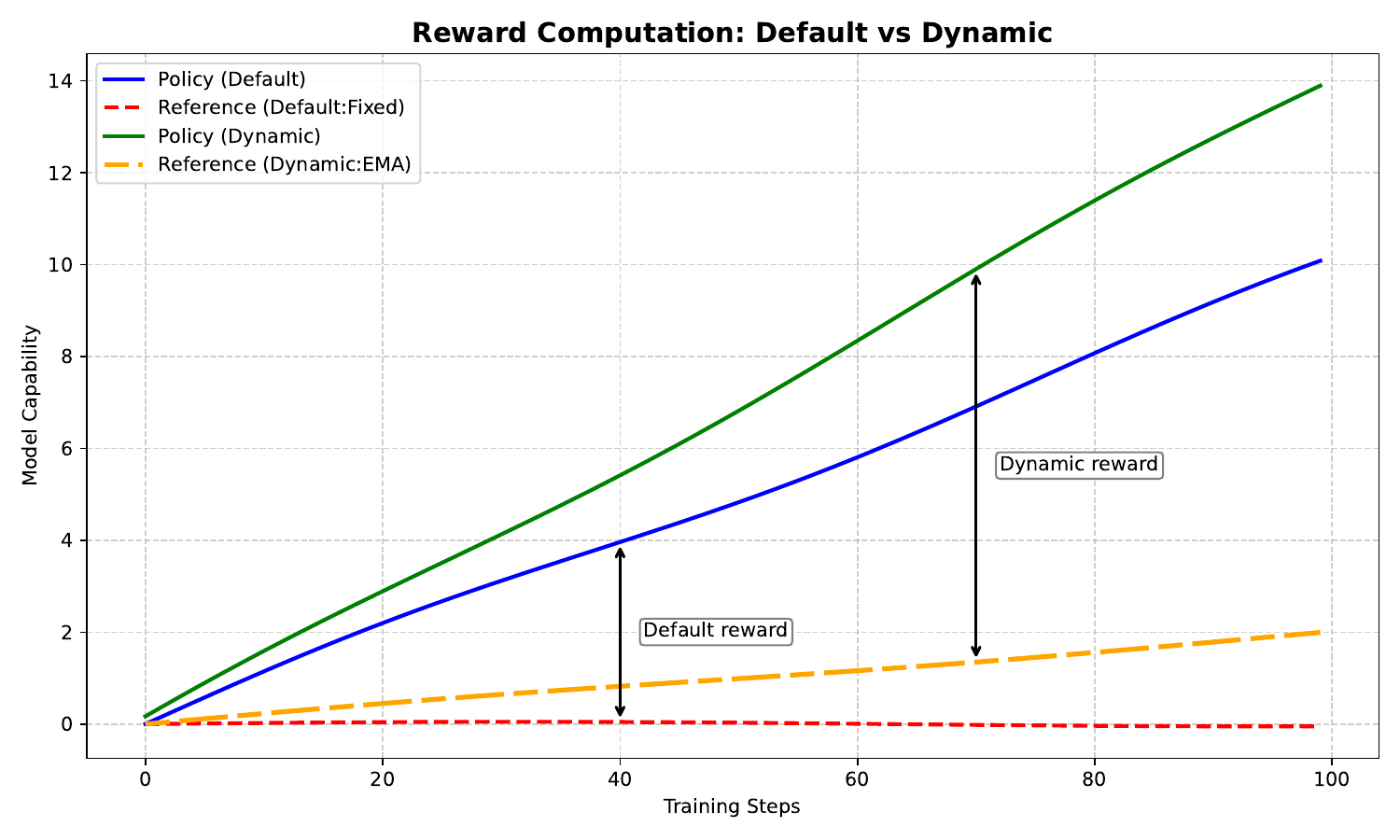}
    \caption{Schematic illustration of reward computation in the baseline and our dynamic framework. In the baseline ~\ref{alg:rl}, rewards are measured as the vertical gap between the policy (blue) and a fixed reference (red). In the dynamic method ~\ref{alg:dy_rl}, rewards are computed against an EMA-updated reference (orange) that evolves smoothly with the policy (green), forming a dynamic baseline. This figure is for conceptual understanding only and does not reflect actual parameter values or training dynamics.}
    \label{fig:dy_rl}
\end{figure}

\begin{algorithm}[htbp]
    \caption{Dynamic Multi-Task Reinforcement Learning from Human Feedback}
    \label{alg:dy_rl}
    \renewcommand{\algorithmicrequire}{\textbf{Dataset:}}
    \renewcommand{\algorithmicensure}{\textbf{Input:}}
    
    \begin{algorithmic}[1]
        \REQUIRE Multi-Task image-condition datasets
        $\{\mathcal{D}_k\}^K_{k=1}$, with data sample probability distribution $\mathcal{P} = \{p_1, p_2, \dots, p_K\}$, task ids $S = \{s_1, s_2, \dots, s_K\}$ and each evaluation dimension $\{\mathcal{E}_k\}^K_{k=1}$. 
        \ENSURE Reference diffusion model $\pi_{ref}$, policy model $\pi_{\theta}$ with parameters $\theta$, unified reward model $r$ with parameters $\phi$, hyperparameters $[t_1, t_2]$ for the generation of evaluate image.

        \STATE Init reference model $\pi_{ref} \leftarrow \pi_{\theta}$
        \FOR {iteration = 1, \dots, N}

        \STATE Sample condition $c$ from the $k$-th dataset $\mathcal{D}_k $with probability $p_k$
        \STATE Sample init noise $\epsilon_1, \epsilon_2$ from normal distribute $\mathcal{N}(0, 1)$
        \STATE Random sample denoise timesteps $t$ from $[t_1, t_2] $  
        \STATE Generate the reference image $x_{ref}$ with full denoise procedure $\pi_{ref}(\epsilon_1,c)$   
        \STATE Generate the evaluate image $x_{\theta}$ with random denoise steps $\pi_\theta(\epsilon_2,c, t)$
        \FOR {$e\in \mathcal{E}_k$}
            \STATE Generate query $q$ with task id $s_k$ and current evalution dimension $e$ as shown in Fig.~\ref{fig:rm_train}
            \STATE Compute RL loss $\mathcal{J}_e(x_\theta,x_{ref},q)$ in Equation~\ref{equ.rl_loss} with reward model $r$
        \ENDFOR
        
        \STATE Updata policy model via gradient ascent:$\pi_{\theta}\leftarrow \pi_{\theta}+\frac{1}{|\mathcal{E}_k|}\nabla_{\pi_{\theta}}\sum\limits_{e \in \mathcal{E}_k}\mathcal{J}_e$
        \STATE $\triangleright$ \textcolor{blue!70!black}{EMA update $\pi_{ref} \leftarrow \tau\pi_{ref} +(1-\tau)\pi_{\theta}$}
        \ENDFOR

        \OUTPUT $\pi_{\theta}, \pi_{ref}$
    \end{algorithmic}
\end{algorithm}